\documentclass[10pt,conference]{IEEEtran}

\usepackage{amsmath}
\usepackage{cite}
\usepackage{amsmath, amssymb, epsfig, amsthm}
\usepackage{mathtools}
\usepackage{mathrsfs}
\usepackage{color}
\usepackage[table]{xcolor}
\usepackage{url}
\usepackage{algorithmicx}
\usepackage{algpseudocode}
\usepackage[ruled]{algorithm2e}
\usepackage{graphicx}
\usepackage{caption}
\usepackage{comment}
\usepackage{enumitem}
\usepackage{color}
\usepackage{subfig}
\usepackage{wrapfig}
\usepackage{bm}

\newtheorem{conjecture}{Conjecture}
\newtheorem{theorem}{Theorem}[section]

\newcommand{\U}{{\mathbf{U}}}
\newcommand{\V}{{\mathbf{V}}}
\newcommand{\X}{{\mathbf{X}}}
\newcommand{\D}{{\mathbf{D}}}
\newcommand{\Z}{{\mathbf{Z}}}
\newcommand{\bW}{{\mathbf{W}}}
\newcommand{\Zopt}{{\mathbf{Z}^*}}
\newcommand{\Wopt}{{\mathbf{W}^*}}
\newcommand{\Id}{{\mathbf{Id}}}

\newcommand{\bSigma}{{\mathbf{\Sigma}}}
\newcommand{\y}{{\mathbf{y}}}
\newcommand{\subscript}[2]{$#1 _ #2$}

\newcommand{\bP}{{\mathbf{P}}}
\newcommand{\E}{{\mathbf{E}}}
\newcommand{\col}[1]{#1_{(\cdot,j)}}
\newcommand{\row}[1]{#1_{(i,\cdot)}}

\newcommand{\entry}[1]{#1_{(i,j)}}

\DeclareMathOperator*{\argmin}{arg\,min}

\begin{document}

\title{Analysis of Fast Alternating Minimization for Structured Dictionary Learning}

\author{\IEEEauthorblockN{Saiprasad~Ravishankar}
\IEEEauthorblockA{Department of Electrical Engineering\\ and Computer Science\\
University of Michigan, Ann Arbor \\
MI, 48109 USA\\
ravisha@umich.edu}
\and
\IEEEauthorblockN{Anna Ma}
\IEEEauthorblockA{Institute of Mathematical Sciences \\
Claremont Graduate University \\
CA, 91711 USA \\
anna.ma@cgu.edu}
\and
\IEEEauthorblockN{Deanna Needell}
\IEEEauthorblockA{Department of Mathematics \\
University of California Los Angeles\\
CA, 90095 USA\\
deanna@math.ucla.edu}
}

\maketitle

\begin{abstract}
Methods exploiting sparsity have been popular in imaging and signal processing applications including compression, denoising, and imaging inverse problems. Data-driven approaches such as dictionary learning enable one to discover complex image features from datasets and provide promising performance over analytical models. Alternating minimization algorithms have been particularly popular in dictionary and transform learning. In this work, we study the properties of alternating minimization for structured (unitary) sparsifying operator learning. While the algorithm converges to the stationary points of the non-convex problem in general, we prove local linear convergence to the underlying generative model under mild assumptions. Our experiments show that the unitary operator learning algorithm is robust to initialization.
\end{abstract}


\section{Introduction} \label{sec1}

Various models of signals and images have been studied in recent years including dictionary models, tensor and manifold models.
Dictionaries that sparsely represent signals are used in applications such as compression, denoising, and medical image reconstruction.
Dictionaries learned from training data sets may outperform analytical models since they are adapted to signals (or signal classes).
The goal of dictionary learning is to find a matrix $\mathbf{D}$ such that an input matrix $\bP$, representing the data set, can be written as $\bP \approx \mathbf{D} \Z$, with $\Z$ denoting the (unknown) sparse representation matrix.

The learning of synthesis dictionary and sparsifying transform models has been studied in several works \cite{elad, Yagh, Mai, barchi1, smith1, bao1, soupsai, sravbrijefraj, sabres, sabres3, saiwen}. 
The convergence of specific learning algorithms has been studied in recent works \cite{spel2b, arora1, yint3, bao1, agra2, soupsai, sbclsTS2}. The learning problems are typically non-convex and some of these works prove convergence to critical points in the problems \cite{bao1, bao2, soupsai}. Others (e.g., \cite{agra1, agra2}) prove recovery of generative models for specific (often computationally expensive) algorithms, but rely on many restrictive assumptions.
A very recent work considers a structured dictionary learning objective showing with high probability that there are no spurious local minimizers, and also provides a specific convergent algorithm \cite{sun1, sun2}.

In this work, we analyze the convergence properties of a structured (unitary) dictionary or transform learning algorithm that involves computationally cheap updates and works well in applications such as image denoising and magnetic resonance image reconstruction \cite{sabres3, sbclsTS2, sravTCI1}.
Our goal is to simultaneously find an $n \times n$ sparsifying transformation matrix $\bW$ and an $n \times N$ sparse coefficients (representation) matrix $\Z$ for training data represented as columns of a given $n \times N$ matrix $\bP$, by solving the following constrained optimization problem:
\begin{equation}
\argmin_{\bW, \Z} \|\bW\bP - \Z \|^2_F  \;\;\; \text{s.t.} \;\; \bW^T\bW = \Id, \, \begin{Vmatrix}
\Z_{(.,j)}
\end{Vmatrix}_{0}\leq s \, \forall j.
\label{eq:opt_program}
\end{equation}
The columns $\Z_{(.,j)}$ of $\Z$ have at most $s$ non-zeros (corresponding to the $\ell_{0}$ ``norm"), where $s$ is a given parameter.
Alternatives to Problem \eqref{eq:opt_program} would replace the column-wise sparsity constraint with a constraint on the sparsity of the entire matrix $\Z$ (i.e., aggregate sparsity), or use a sparsity penalty (e.g., $\ell_{p}$ penalties with $0\leq p \leq  1$).
Problem \eqref{eq:opt_program} also corresponds to learning a (synthesis) dictionary $\bW^{T}$ for representing the data $\bP$ as $\bW^{T}\Z$.

Optimizing Problem \eqref{eq:opt_program} by alternating between updating $\bW$ (operator update step) and $\Z$ (sparse coding step) would generate the following updates. The $t$th $\Z$ update is given as ${\Z}_{(.,j)}^{t} = H_{s}({\bW}^{t-1} \bP_{(.,j)})$ $\forall$ $j$, where the thresholding operator $H_{s}(\cdot)$ zeros out all but the $s$ largest magnitude elements of a vector (leaving the $s$ entries unchanged). The subsequent $\bW$ update is based on the full singular value decomposition (SVD) of ${\Z}^t \bP^{T} \triangleq \V \mathbf{\Sigma} \U^{T}$ with ${\bW}^{t}=\V \U^{T}$. The method is shown in Algorithm~\ref{alg:alternate}.

Recent works have shown convergence of Algorithm~\ref{alg:alternate} or its variants to critical points in the equivalent unconstrained problems \cite{sbclsTS2, sabsam, bao22}.
Here, we further prove local linear convergence of the method to the underlying generative (data) model under mild assumptions that depend on properties of the underlying/generating sparse coefficients.
Our experiments show that the method is also robust or insensitive to initialization in practice.

\begin{algorithm}
 \caption{Alternating Optimization for \eqref{eq:opt_program}} \label{alg:alternate}
\begin{algorithmic}
		\State \textbf{Input: } Training data matrix $\bP$, maximum iteration \\ count $L$, sparsity $s$
		\State \textbf{Output: }$\bW^L$, $\Z^L$ 
		\State \textbf{Initialize: } $\bW^0$ and $t=1$
	   \State \For{$t \leq L$}{
		   \State $\Z_{(.,j)}^t = H_{s}\begin{pmatrix}\bW^{t-1} \bP_{(.,j)} \end{pmatrix}$ $\forall$ $j$ 
		   \State $\bP {\Z^t}^{T}$=$\U^t \bSigma^t {\V^t}^T$  
		  \State $\bW^t = \V^t{\U^t}^T$
		  \State $t = t+1$
		  }
\end{algorithmic}
\end{algorithm}

\section{Convergence Analysis} \label{sec2}

The main contribution of this work is the local convergence analysis of Algorithm~\ref{alg:alternate}. In particular, we show that under mild assumptions, the iterates in Algorithm~ \ref{alg:alternate} converge linearly to the underlying (generating) data model.

\subsection{Notation} \label{sec2:not}
In the remainder of this work, we adopt the following notation. The unitary transformation matrix, sparse representation matrix, and training data matrix are denoted as $\bW \in \mathbb{R}^{n \times n}$, $\Z \in \mathbb{R}^{n \times N}$, and $\bP \in \mathbb{R}^{n \times N}$, respectively. For a matrix $\X$, we denote its $j$th column, $i$th row, and entry $(i,j)$ by $\col{\X}$, $\row{\X}$, and $\entry{\X}$ respectively. 
The $t$th iterate or approximation in the algorithm is denoted using $(\cdot)^t$, with a lower-case $t$, and $(\cdot)^T$ denotes the transpose.
The function $S(\y)$ returns the support (i.e., set of indexes) of a vector $\y \in \mathbb{R}^{n}$, or the locations of the nonzero entries in $\y$. 
Matrix $\D_k$ denotes an $n \times n$ diagonal matrix of ones and a zero at location $(k,k)$. Additionally, $\tilde{\D}_k$ denotes an $N \times N$ diagonal matrix that has ones at entries $(i,i)$ for $i \in S(\Zopt_{(k,\cdot)})$ and zeros elsewhere, and matrix $\Zopt$ is defined in Section \ref{sec:assump} (see assumption $(A_1)$). 
The Frobenious norm, denoted $\| \X \|_F^2$, is the the sum of squared elements of $\X$, and $\| \X \|_2$ denotes the spectral norm. Lastly, $\Id$ denotes the appropriately sized identity matrix.

\subsection{Assumptions} \label{sec:assump}
Before presenting our main results, we will briefly discuss our assumptions and explain their implications:
\begin{enumerate}[label=(\subscript{A}{\arabic*})]
\itemsep0em 
\item \textbf{Generative model:} There exists a $\Z^*$ and unitary $\bW^*$ such that  $\bW^* \bP = \Z^*$, and $\left \| \bP \right \|_{2}=1$ (normalized).% A1
\item \textbf{Sparsity:} The columns of $\Z^*$ are $s-$sparse, i.e., $\| \col{\Z}^* \|_0 \leq s$ $\forall j$. % A2
\item \textbf{Spectral property:} The underlying $\Zopt$ satisfies the bound $\kappa^{4}\begin{pmatrix}
\Zopt
\end{pmatrix}\max_{1\leq k \leq n} \| \D_k \Zopt \Zopt^T\Zopt \tilde{\D}_k \|_2 < 1$, where $\kappa(\cdot)$ denotes the condition number (ratio of largest to smallest singular value).% A3
\item \textbf{Initialization:} $\| \bW^0 - \bW^* \|_{F} \leq \epsilon$ for an appropriate sufficiently small $\epsilon >0$. \footnote{Although we do not specify the best (largest permissible) $\epsilon$ explicitly, $\epsilon < \frac{1}{2} \min_j \beta \left( \frac{\col{\Z}^*}{\|\col{\Z}^*\|_{2}} \right)$ with $\beta(\cdot)$ denoting the smallest nonzero magnitude in a vector, will arise in one of our proof steps. The actual permissible $\epsilon$ is also dictated as per (convergence of) Taylor series expansions discussed in the proof.} %A4
\end{enumerate}

The generative model assumption simply states that there exists an underlying transform and representation matrix for the data set $\bP$. So we would like to investigate if Algorithm~\ref{alg:alternate} can find such underlying models (minimizers in \eqref{eq:opt_program}).
Assumption $(A_2)$ states that the columns of $\Zopt$ have at most $s$ nonzeros.
We assume that the coefficients are ``structured" in assumption $(A_3)$, satisfying a spectral property, which will be used to establish our theorems. Later we discuss a conjecture that states that this property holds for a specific probabilistic model.
Assumption $(A_4)$ states that the initial sparsifying transform is sufficiently close to the solution $\bW^*$. This assumption also simplifies our proof and has been made in other works such as \cite{agra1, agra2}, which address the issue of specific (good) initialization separately from  the main algorithm. In this work, we empirically show the effect of general initializations in Section \ref{sec3}.

\subsection{Main Results and Proofs}

We state the convergence results in Theorems~\ref{thm:main1} and \ref{thm:main2}.
Theorem~\ref{thm:main1} establishes that Algorithm~\ref{alg:alternate} converges to the underlying generative model under the assumptions discussed in Section~\ref{sec:assump}. It also makes an additional assumption on $\Zopt$ that $\Zopt\Zopt^T = \Id$, which simplifies assumption $(A_3)$.
Theorem~\ref{thm:main2} presents the more general result based on Assumption $(A_3)$. 
We only include the (simpler) proof of Theorem~\ref{thm:main1}, while the full details of the general Theorem \ref{thm:main2}'s proof can be found in \cite{saiannadeanna1}. 
Following Theorem~\ref{thm:main2}, Conjecture~\ref{conj} states that Assumption $(A_3)$ holds
under a commonly used probabilistic assumption on the sparse representation matrix $\Zopt$.

\begin{theorem} Under Assumptions $(A_1) - (A_4)$ and assuming $\Zopt \Zopt^{T} =\Id$, the Frobenius error between the iterates generated by Algorithm~\ref{alg:alternate} and the underlying generative model in Assumption $(A_1)$ is bounded as follows:
\begin{equation}
\| \Z^t - \Zopt \|_F  \leq  q^{t-1} \epsilon, \;\; \| \bW^t - \Wopt \|_F  \leq  q^t \epsilon, 
\end{equation}
where $q \triangleq \max_{1 \leq k \leq n} \| \D_k \Zopt \tilde{\D}_k \|_2$ and $\epsilon$ is fixed based on the initialization.
\label{thm:main1}
\end{theorem}

Since $\Zopt \Zopt^{T} =\Id$, by Assumption $(A_3)$ it follows that $q<1$ above, and thus the Theorem establishes that the iterates converge at a linear rate to the underlying  generative model. 

In the following, we prove Theorem~\ref{thm:main1} using induction on the approximation error of iterates with respect to $\Zopt$ and $\bW^*$. 
Let the series $ \{ \E^t \}$ and $\{ \mathbf{ \Delta}^t \}$ be defined as
\begin{align}
\E^t &:=\bW^t - \bW^*,  \label{eq:deltaw}\\
\mathbf{\Delta}^t &:= \Z^t -\Zopt. \label{eq:deltaz}
\end{align}
By Assumption $(A_4)$, $\| \E^0 \|_F \leq \epsilon$. We first provide a proof for the base case of $t=1$. This proof involves two main parts. First, we show that the error between $\Z^1$ and $\Zopt$ is bounded (in norm) by $\epsilon$. Then, we show that $\| \bW^1 - \Wopt \|_F  \leq  q \epsilon$, where $q$ is iteration-independent.

For the first part, we use Assumptions $(A_1)$, $(A_2)$, and $(A_4)$. Assumption $(A_4)$ ensures that a superset of the support of $\Zopt$ is recovered in the first iteration. In particular, each  column of the sparse coefficients matrix $\Z^{1}$ in Algorithm~\ref{alg:alternate} satisfies
\begin{align}
\col{\Z}^1 \stackrel{}{=} H_s( & \bW^{0}  \col{\bP}) \stackrel{(Eq. \ref{eq:deltaw})}{=} H_s(\Wopt \col{\bP} + \E^{0} \col{\bP}) \nonumber  \\
&\stackrel{(A_1)}{=} H_s(\col{\Z}^* + \E^{0} \col{\bP}) \nonumber \\ 
&\stackrel{(A_4)}{=} \col{\Z}^* + \mathbf{\Gamma}_j^1 \E^{0} \col{\bP}, \label{eq:zi}
\end{align}
where $\mathbf{\Gamma}_j^1$ is a diagonal matrix with a one in the $(i,i)$th entry if $i \in S(\col{\Z}^1)$ and zero otherwise and $\E^{0}$ is as defined in \eqref{eq:deltaw}.
The last equality in \eqref{eq:zi} follows from the fact that the support of $\col{\Z}^1$ includes that of $\col{\Z}^*$ for small $\epsilon$. In particular, since $\begin{Vmatrix}
\col{\bP}
\end{Vmatrix}_{2} = \begin{Vmatrix}
\col{\Z}^*
\end{Vmatrix}_{2}$, we have 
\begin{align*}
\begin{Vmatrix}
\E^{0} \col{\bP}
\end{Vmatrix}_{\infty}\leq \begin{Vmatrix}
\E^{0} \col{\bP}
\end{Vmatrix}_{2} \leq \begin{Vmatrix}
\E^{0}
\end{Vmatrix}_{F} \begin{Vmatrix}
\col{\Z}^*
\end{Vmatrix}_{2}.
\end{align*}
Therefore, whenever $\| \E^0 \|_F \leq \epsilon < \frac{1}{2} \min_j \beta \left( \frac{\col{\Z}^*}{\|\col{\Z}^*\|_{2}} \right)$ with $\beta(\cdot)$ denoting the smallest nonzero magnitude in a vector, the support of $\col{\Z}^1$ includes that of $\col{\Z}^*$ (i.e., the entries of the perturbation term $\E^{0} \col{\bP}$ are not large enough to change the support).
The following results then hold:
\begin{align*}
\|\Z^1 - \Zopt \|_F^{2} &\stackrel{(Eq. \ref{eq:zi})}{=} \|[\mathbf{\Gamma}_1^1 \E^{0} \bP_{(\cdot,1)} , ... , \mathbf{\Gamma}_N^1 \E^{0} \bP_{(\cdot,N)}] \|^2_F \\
& \stackrel{(i)}{\leq} \| \E^{0} \bP\|_F^{2} \stackrel{(ii)}{\leq} \| \E^{0}\|_F^2 \| \bP \|^2_2 \stackrel{(A_1)}{=} \| \E^{0}\|_F^2.
\end{align*}
Step $(i)$ follows by definition of $\mathbf{\Gamma}_j^1$; step $(ii)$ holds for the Frobenius norm of a product of matrices; and since $\| \bP\|_2^2 = 1$, the last equality holds. 
Therefore, we can conclude that
\begin{equation}
\| \Z^1 - \Zopt \|_F \leq \| \E^{0}\|_F \stackrel{(A_4)}{\leq} \epsilon.
\label{eq:errorZ}
\end{equation}

Next, we analyze the quality of the updated transform $\bW^1$. To bound $\| \bW^1 - \Wopt \|_F $ by $q\epsilon$, we rely on Taylor Series expansions of the matrix inverse and positive-definite square root functions. %which have simple forms under the assumption $\Zopt \Zopt^{T}=\Id$.
Denote the full SVD of $\Zopt {\Z^1}^T$ as $\U_z^1 \bSigma_z^1 {\V_z^1}^T$. Then, from Assumption $(A_1)$ and Algorithm~\ref{alg:alternate}, we have,
\begin{align*}
\Wopt^{T} \Zopt {\Z^1}^T \stackrel{(A_1)}{=}  \bP {\Z^1}^T = \U^1 \bSigma^1 {\V^1}^T, \;\; \bW^1 = \V^1{\U^1}^T.
\end{align*}
Then, $\bW^1$ is expressed in terms of the SVD of $\Zopt {\Z^1}^T$ as 
\begin{equation}
\bW^1 = \V_z^1{\U_z^1}^T \Wopt. 
\label{eq:w1}
\end{equation}

Using \eqref{eq:w1}, the error between $\bW^1$ and $\Wopt$ satisfies
\begin{align}
\| & \bW^1 - \Wopt \|_F  = \| \V_z^1{\U_z^1}^T \Wopt - \Wopt \|_F \nonumber\\
& = \| (\V_z^1{\U_z^1}^T  - \Id) \Wopt \|_F = \| \V_z^1{\U_z^1}^T  - \Id\|_F, \label{eq:errorw1}
\end{align}
where the matrix $\V_z^1{\U_z^1}^T$ can be further rewritten as follows:
\begin{align}
\V_z^1{\U_z^1}^T &= \V_z^1 \begin{pmatrix}
\bSigma^{1}_z
\end{pmatrix}^{-1} {\U_z^1}^T\U_z^1 \bSigma^{1}_z {\U_z^1}^T \nonumber \\
& = \underbrace{(\Zopt {\Z^1}^T)^{-1}}_{(a)} \underbrace{(\Zopt {\Z^1}^T {\Z^1} \Zopt^T )^{\frac{1}{2}}}_{(b)}.
\label{eq:vu}
\end{align}
It is easy to show that since $\Zopt \Zopt^T =\Id$, $\Zopt {\Z^1}^T$ is invertible for all $\epsilon < 1$ (sufficient condition).

Using \eqref{eq:deltaz} and the assumption $\Zopt \Zopt^{T} = \Id$, the Taylor Series expansions for the matrix inverse and positive-definite square root in \eqref{eq:vu},
can be written as
\begin{align*}
(a) &= (\Zopt {\Z^1}^T)^{-1} = (\Id + \Zopt {\mathbf{\Delta}^1}^T)^{-1} \\
&= \Id - \Zopt {\mathbf{\Delta}^1}^T + O ((\mathbf{\Delta}^1)^2)\\
(b) &= (\Zopt {\Z^1}^T {\Z^1} \Zopt^T )^{\frac{1}{2}} \\
& = \Id + \frac{1}{2}(\Zopt {\mathbf{\Delta}^1}^T + \mathbf{\Delta}^1 \Zopt^T) + O((\mathbf{\Delta}^1)^2).
\end{align*}
Therefore we can rewrite \eqref{eq:vu} as
\begin{align*}
\V_z^1{\U_z^1}^T & = (a)(b) \nonumber \\
&= \Id + \frac{1}{2}(\mathbf{\Delta}^1 \Zopt^T - \Zopt {\mathbf{\Delta}^1}^T ) + O((\mathbf{\Delta}^1)^2),
\end{align*}
where $O((\mathbf{\Delta}^1)^2)$ denotes corresponding higher order series terms, and is bounded in norm by $C \begin{Vmatrix}
\mathbf{\Delta}^{1}
\end{Vmatrix}^{2}$ for some constant $C$ (independent of the iterates).

Substituting the above expressions in \eqref{eq:errorw1}, the error between the first transform iterate $\bW^1$ and $\Wopt$ is bounded as
\begin{align}
\nonumber \|\bW^1 - \Wopt \|_F &\stackrel{( Eq. \ref{eq:errorw1})}{=} \| \V_z^1{\U_z^1}^T  - \Id\|_F \\
& \approx \frac{1}{2} \| \mathbf{\Delta}^1 \Zopt^T - \Zopt {\mathbf{\Delta}^1}^T \|_F. \label{upb1}
\end{align}
The approximation error in \eqref{upb1} is bounded in norm by $C \epsilon^2$, which is negligible for small $\epsilon$. So we only bound the (dominant) term $0.5 \| \mathbf{\Delta}^1 \Zopt^T - \Zopt {\mathbf{\Delta}^1}^T \|_F$. 
Since the matrix $\mathbf{\Delta}^1 \Zopt^T - \Zopt {\mathbf{\Delta}^1}^T$ clearly has a zero diagonal, we have the following inequalities:

\begin{align*}
&\|\bW^1 - \Wopt \|_F  \approx \frac{1}{2} \| \mathbf{\Delta}^1 \Zopt^T - \Zopt {\mathbf{\Delta}^1}^T \|_F \\
& \hspace{-0.06in} \leq \sqrt{ \sum_{k=1}^{n} \|  \D_k \Zopt \tilde{\D}_k {\mathbf{\Delta}_{(k,\cdot)}^1}^T \|_2^2 }  \leq  \sqrt{ \sum_{k=1}^{n} \|  \D_k \Zopt \tilde{\D}_k \|_2^2 \| \mathbf{\Delta}_{(k,\cdot)}^1 \|_2^2} \\
& \leq \max_k   \|  \D_k \Zopt \tilde{\D}_k\|_2 \sqrt{ \sum_{k=1}^{n} \| \mathbf{\Delta}_{(k,\cdot)}^1 \|_2^2  }  =  q \| \Z^1 - \Zopt \|_F \stackrel{\eqref{eq:errorZ}}{\leq} q\epsilon,
\end{align*}
where $q \triangleq   \max_k  \|  \D_k \Zopt \tilde{\D}_k\|_2$. 

Thus, we have shown the results for the $t=1$ case.
We complete the proof of Theorem~\ref{thm:main1} by observing that for each subsequent iteration $t = \tau+1$, the same steps as above can be repeated along with the induction hypothesis (IH) to show that 
\begin{align*}
\|\Z^{\tau+1} - \Zopt \|_F &= \| \mathbf{\Delta}^{\tau+1} \|_F \leq \| \E^{\tau}\|_F \\
& = \| \bW^{\tau} - \Wopt \|_F  \stackrel{(IH)}{\leq} q^\tau \epsilon\\
\| \bW^{\tau+1} - \Wopt \|_F &\leq q \| \Z^{\tau + 1} - \Zopt \|_F  \leq q(q^{\tau} \epsilon). \;\;\;\;\;\;\;\;\; \blacksquare
\end{align*} 

The next result generalizes Theorem~\ref{thm:main1} by removing the assumption $\Zopt \Zopt^{T} = \Id$.

\begin{theorem} Under Assumptions $(A_1)-(A_4)$, the iterates in Algorithm~ \ref{alg:alternate} converge linearly to the underlying generative model in Assumption $(A_1)$, i.e., the Frobenius error between the iterates and the generative model satisfies
\begin{equation}
\| \Z^t - \Zopt \|_F  \leq  q^{t-1} \epsilon, \;\; \| \bW^t - \Wopt \|_F  \leq  q^t \epsilon, 
\end{equation}
where $q \triangleq \kappa^{4}\begin{pmatrix}
\Zopt
\end{pmatrix}\max_{1\leq k \leq n} \| \D_k \Zopt \Zopt^T\Zopt \tilde{\D}_k \|_2 < 1$ and $\epsilon$ is fixed based on the initialization. 
\label{thm:main2}
\end{theorem}

Note that dropping the unit spectral norm (normalization) condition on $\bP$ in Assumption $(A_1)$ does not affect the $\| \bW^t - \Wopt \|_F$ bound in Theorem~\ref{thm:main2} and only creates a scaling in the $\| \Z^t - \Zopt \|_F$ bound, where the $\epsilon$ gets replaced by $\| \bP \|_{2} \epsilon$.

\begin{conjecture} Suppose the locations of the $s$ nonzeros in each column of $\Zopt$ is chosen uniformly at random, and the non-zero entries are i.i.d. as $\entry{\Zopt} \sim \mathcal{N}(0,\frac{n}{sN})$. Then, for fixed, small $\frac{s}{\sqrt{n}}$, $q \triangleq \kappa^{4}\begin{pmatrix}
\Zopt
\end{pmatrix}\max_{1\leq k \leq n} \| \D_k \Zopt \Zopt^T\Zopt \tilde{\D}_k \|_2 < 1$ for large enough $N$.
\label{conj}
\end{conjecture}

Conjecture~\ref{conj} thus states that under the assumed probabilistic model for $\Zopt$, when $N$ is large enough or there is sufficient training data (or columns of $\bP$), then Algorithm~\ref{alg:alternate} is assured to have rapid local linear iterate convergence to the underlying generative model. 
This conjecture can be empirically verified through simulations and the numerical results supporting it can be found in \cite{saiannadeanna1}. 
The experiments presented in this paper will focus on illustrating the local convergence of Algorithm~\ref{alg:alternate} and its robustness to initialization.

\vspace{0.1in}
\begin{figure}[!t]
\centering
\includegraphics[width=.45\textwidth]{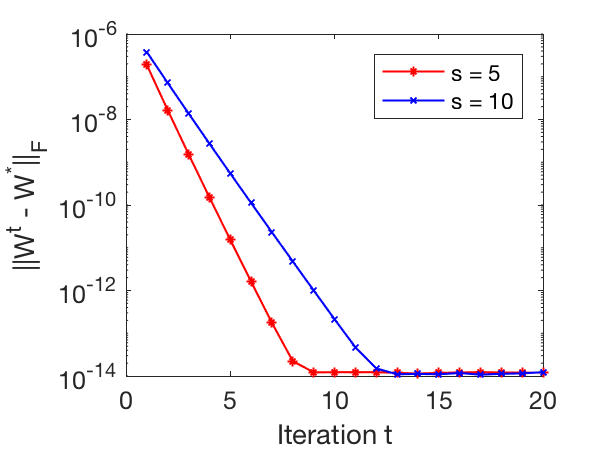}
\caption{The performance of Algorithm \ref{alg:alternate} for recovering $\Wopt$ for $s=5$ and $s=10$.}
\label{fig:errorW}
\end{figure}

\section{Experiments} \label{sec3}

In this section, we show numerical experiments in support of our analytical conclusions. We also provide results further illustrating the robustness of the algorithm to initializations.

In our experiments, we generated the training data using randomly generated $\Wopt$ and $\Zopt$, with $n = 50$, $N = 10000$, and $s=\{ 5, \, 10 \}$.
The transform $\Wopt$ is generated in each case by applying Matlab's \texttt{orth()} function on a standard Gaussian matrix.
The representation matrix $\Zopt$ is generated for each $s$ as described in Conjecture~\ref{conj}, i.e., the support of each column of $\Zopt$ is chosen uniformly at random and the nonzero entries are drawn i.i.d. from a Gaussian distribution with mean zero and variance $n/sN$.

In the first experiment, the initial $\bW^0$ in Algorithm~\ref{alg:alternate} is chosen to satisfy $\| \bW^0 - \Wopt \|_F \leq \epsilon$ with $\epsilon=0.49 \min_j \beta \left( \frac{\col{\Z}^*}{\|\col{\Z}^*\|_{2}} \right)$ (see \eqref{eq:zi}).
Fig.~\ref{fig:errorW} shows the behavior of the Frobenious error between $\bW^t$ and $\Wopt$ in the algorithm. %Algorithm~\ref{fig:errorW}. 
The observed (linear) convergence of the iterates to the generative operator $\Wopt$
is in accordance with Theorem~\ref{thm:main2} 
and our Conjecture~\ref{conj}.

\begin{figure}[!t]
\centering
\includegraphics[width=.4\textwidth]{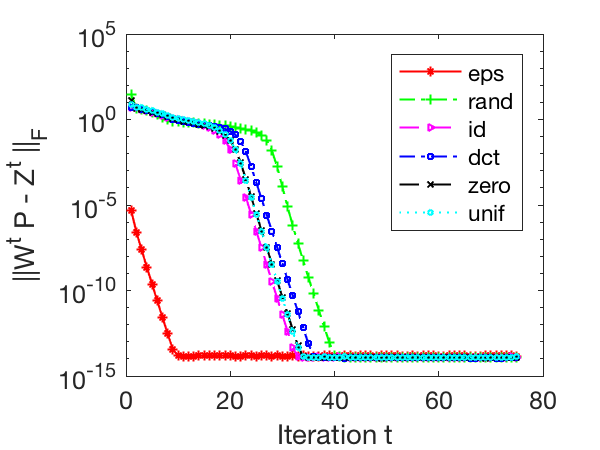}\\
\includegraphics[width=.4\textwidth]{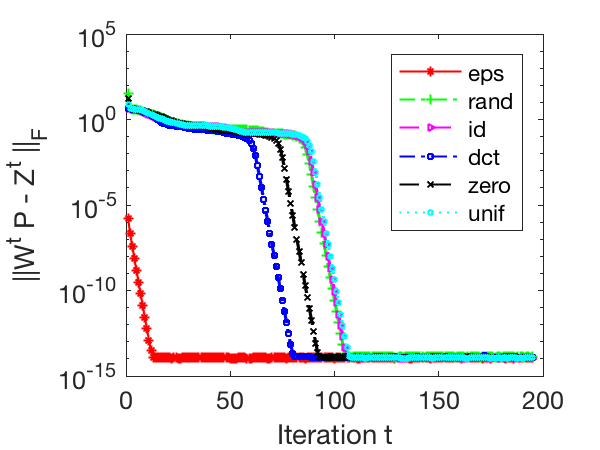}
\caption{The performance of Algorithm \ref{alg:alternate} with various initializations for $s=5$ (top) and $s=10$ (bottom).}
\label{fig:init}
\end{figure}

Next, we study the performance of Algorithm~\ref{alg:alternate} with different initializations for $n = 50$, $N = 10000$, and $s=\{ 5, \, 10 \}$. 
Fig.~\ref{fig:init} shows the objective function in Problem~\eqref{eq:opt_program} over the algorithm iterations. Since the training data satisfy Assumptions $(A_1)$ and $(A_2)$, the minimum objective value in \eqref{eq:opt_program} is $0$.
Six different types of initializations are considered.
The first, labeled `eps', denotes an initialization as in Fig.~\ref{fig:errorW} with $\epsilon=0.49 \min_j \beta \left( \frac{\col{\Z}^*}{\|\col{\Z}^*\|_{2}} \right)$. The other initializations are as follows: entries of $\bW^0$ drawn i.i.d. from a standard Gaussian distribution (labeled `rand'); an $n \times n$ identity matrix $\bW^0$ labeled `id'; a discrete cosine transform (DCT) initialization labeled `dct'; entries of $\bW^0$ drawn i.i.d. from a uniform distribution ranging from 0 to 1 (labeled `unif'); and $\bW^0 = \textbf{0}^{n \times n}$ labeled `zero'. 
For more general initializations (other than `eps'), we see that the behavior of Algorithm~\ref{alg:alternate} is split into two phases. In the first phase, the iterates slowly decrease the objective. 
When the iterates are close enough to a solution, the second phase occurs and during this phase, Algorithm~\ref{alg:alternate} enjoys rapid convergence (towards 0). 
Note that the objective's convergence rate in the second phase is comparable to that of the `eps' case. 
The behavior of Algorithm~\ref{alg:alternate} is similar for $s=5$ and $s=10$, with the latter case taking more iterations to enter the second phase of convergence. This makes sense since there are more variables to learn for larger $s$.

\section{Conclusion} \label{sec4}
This work presented an analysis of a fast alternating minimization algorithm for unitary sparsifying operator learning. We proved local linear convergence of the algorithm to the underlying generative model under mild assumptions. Numerical experiments illustrated this local convergence behavior, and demonstrated that the algorithm is robust to initialization in practice. The full version of this work, including the proof of Theorem~\ref{thm:main2} and numerical results that support Conjecture~\ref{conj} can be found in \cite{saiannadeanna1}. A theoretical analysis of the algorithm's robustness observed in Fig.~\ref{fig:init} is left for future work. 

\section*{Acknowledgments}
Saiprasad Ravishankar was supported in part by the following grants: NSF grant CCF-1320953,
ONR grant N00014-15-1-2141, DARPA Young Faculty Award D14AP00086, ARO MURI grants W911NF-11-1-0391 and 2015-05174-05, NIH grants R01 EB023618 and U01 EB018753, and a UM-SJTU seed grant. Anna Ma and Deanna Needell were supported by the NSF DMS \#1440140 (while they were in residence at the Mathematical Science Research Institute in Berkeley, California, during the Fall 2017 semester), NSF CAREER DMS \#1348721, and the NSF BIGDATA DMS \#1740325.

\bibliographystyle{IEEEbib}
\bibliography{ita2018v3}

\end{document}